\documentclass[
]{ceurart}

\usepackage{listings}
\usepackage{multirow}

\usepackage{graphicx}
\usepackage[table,xcdraw]{xcolor}
\usepackage[normalem]{ulem}
\useunder{\uline}{\ul}{}

\begin{document}

\copyrightyear{2025}
\copyrightclause{Copyright for this paper by its authors.
  Use permitted under Creative Commons License Attribution 4.0
  International (CC BY 4.0).}

\conference{IberLEF 2025, September 2025, Zaragoza, Spain} 

\title{Tackling a Challenging Corpus for Early Detection of Gambling Disorder: UNSL at MentalRiskES 2025}



\author[1,2]{Horacio Thompson}[
orcid=0009-0009-2103-2275,
email=hjthompson@unsl.edu.ar,
]\cormark[1]

\author[1]{Marcelo Errecalde}[%
email=merreca@unsl.edu.ar,
]

\address[1]{Universidad Nacional de San Luis (UNSL), Ejército de Los Andes 950, San Luis, C.P. 5700, Argentina}
\address[2]{Consejo Nacional de Investigaciones Científicas y Técnicas (CONICET), San Luis, Argentina}

\cortext[1]{Corresponding author.}

\begin{abstract}
  Gambling disorder is a complex behavioral addiction that is challenging to understand and address, with severe physical, psychological, and social consequences. Early Risk Detection (ERD) on the Web has become a key task in the scientific community for identifying early signs of mental health behaviors based on social media activity. This work presents our participation in the MentalRiskES 2025 challenge, specifically in Task 1, aimed at classifying users at high or low risk of developing a gambling-related disorder. We proposed three methods based on a \emph{CPI+DMC approach}, addressing predictive effectiveness and decision-making speed as independent objectives. The components were implemented using the SS3, BERT with extended vocabulary, and SBERT models, followed by decision policies based on historical user analysis. Although it was a challenging corpus, two of our proposals achieved the top two positions in the official results, performing notably in decision metrics. Further analysis revealed some difficulty in distinguishing between users at high and low risk, reinforcing the need to explore strategies to improve data interpretation and quality, and to promote more transparent and reliable ERD systems for mental disorders.
\end{abstract}


\maketitle

\section{Introduction}
According to the World Health Organization, an estimated 1.2\% of the adult population suffers from a gambling disorder, with the risk growing even for young people and children. In recent years, this disorder has been recognized as a behavioral addiction, prompting the Diagnostic and Statistical Manual of Mental Disorders (DSM-5) and the International Classification of Diseases (ICD-11) to reclassify it alongside substance-related disorders \cite{kim2019review}, replacing the term \emph{pathological gambling} with \emph{gambling disorder}. This condition encompasses a wide spectrum of physical, psychological, and social consequences \cite{wohr2021perception}, including high substance use \cite{browne2020framework}, symptoms of anxiety, depression, stress, and impulsivity \cite{bargeron2017psychosocial, de2020relationship, wu2018prevalence}, as well as work-related and financial conflicts, relationship deterioration, and criminal behavior \cite{browne2016assessing}. Technological advancements and widespread access to digital platforms have contributed to the increasing prevalence of this disorder \cite{nunez2025effectiveness}, alongside other behaviors such as compulsive shopping and problematic social media use \cite{mestre2025association}. Furthermore, numerous studies have highlighted the challenges in establishing precise criteria and consistent methods for estimating the prevalence of gambling disorder due to the diversity of assessment tools, risk factors, and controversies surrounding the validity of diagnostic criteria \cite{gabellini2023prevalence, allami2021meta, rash2016review}.

In this context, Early Risk Detection (ERD) on the Web has become a significant research area in recent years, aiming to identify users who exhibit signs of developing a mental health condition as early as possible. Initiatives such as MentalRiskES have fostered research on ERD in Spanish \cite{marmol2023, marmol2024overview}, while CLEF eRisk has promoted similar efforts primarily in English \cite{parapar2024}. Our research group has actively participated in these challenges, addressing the detection of depression and eating disorders in the Spanish language \cite{thompson2023early}, as well as the detection of depression \cite{loyola2021unsl}, pathological gambling \cite{loyola2022unsl, thompson2023strategies}, and anorexia \cite{thompson2024time} in English. In the MentalRiskES 2025 edition \cite{MentalRiskES2025, iberlef2025overview}, a challenge focused on early detection of gambling disorder was proposed, consisting of two tasks: binary classification (Task 1), aimed at identifying users at high (positive) or low (negative) risk; and multiclass classification (Task 2), designed to distinguish the specific type of addiction associated with the disorder, such as \emph{Betting}, \emph{Online Gaming}, \emph{Trading}, and \emph{Lootboxes}. The challenge was conducted in two phases: a training stage, using a labeled corpus provided by the Organizers, and an online evaluation stage, where teams analyzed user posts progressively while interacting with a server in an early environment. Our team participated in Task 1, presenting three proposals based on a \emph{CPI+DMC approach} \cite{loyola2018learning}. This approach conceptualizes ERD as a multi-objective problem, where the goal is to optimize classification effectiveness and decision-making speed independently.
It consists of two components: a \emph{Classification with Partial Information} (CPI) model that processes user content progressively and a policy for \emph{Deciding the Moment of Classification} (DMC) that determines when to make a final decision based on the accumulated evidence. 
While alternatives exist that simultaneously address both objectives \cite{thompson2024temporal}, we opted for a modular design due to the complexity of the problem. For the CPI component, we implemented three different models: SS3, BERT with extended vocabulary, and SBERT. For the DMC component, we designed decision policies that evaluate users based on historical analysis. 

According to the official ranking released by the Organizers, two of our proposals achieved first and second place, with remarkable results in the Macro F$_1$ score and other decision-making metrics. A detailed analysis of the results highlighted the inherent complexity of the task: the differentiation between users at high and low risk is subtle and difficult to define, posing a challenge for both the proposed models and potential human evaluators. The structure of the paper is as follows: Section \ref{sec:corpus} presents details of the corpus and a preliminary analysis of the data; 
Section \ref{sec:methodology} describes the methodology adopted and the models used; Section \ref{sec:results} discusses the results obtained; and Section \ref{sec:conclusion} offers conclusions and future work.

\section{Corpus}
\label{sec:corpus}
To address Task 1, the Organizers developed a corpus \cite{alvarez2025precom} divided into three parts, deployed in the different phases of the challenge (Table \ref{tab:datasets}). The Train and Trial sets were provided for model training and server connection testing, while the Test set was reserved for the final evaluation of participating teams. Each set presents a balanced class distribution, with a similar mean number of posts per user (around 60) and a relatively short post length, averaging fewer than 10 words per post, although some posts are considerably longer. Additionally, users come from the Telegram and Twitch platforms, with a balanced distribution across classes. 

User information is organized at the post level and includes structured metadata such as message ID, round number, user pseudonym, text content, date, and origin platform. For example, a post from user subject1 is represented as: \{id\_message: 123, round: 1, nick: "subject1", message: "...", date: "2021-01-06 04:02:48+01:00", platform: "Telegram"\}. Each user was labeled by the Organizers with a binary class (high or low risk) based on their complete posting history, considering that all users show some level of gambling behavior, though their risk levels vary.

\begin{table}[ht!]
\caption{Overview of the three subsets of the corpus used in Task 1. The number of users (Total, Positive, and Negative) is reported, along with statistics on posts per user and words per post (Mean, Minimum, and Maximum). The distribution of users across platforms (Telegram and Twitch) is also indicated.}
\begin{tabular}{|l|ccc|ccc|ccc|cccc|}
\hline
\multicolumn{1}{|c|}{\multirow{3}{*}{\textbf{Corpus}}} & \multicolumn{3}{c|}{\textbf{\#Users}}                                                                                                     & \multicolumn{3}{c|}{\textbf{\#Posts per user}}                                                                                           & \multicolumn{3}{c|}{\textbf{\#Words per post}}                                                                                           & \multicolumn{4}{c|}{\textbf{Platform}}                                                                                   \\ \cline{2-14} 
\multicolumn{1}{|c|}{}                                 & \multicolumn{1}{c|}{\multirow{2}{*}{\textbf{Total}}} & \multicolumn{1}{c|}{\multirow{2}{*}{\textbf{Pos}}} & \multirow{2}{*}{\textbf{Neg}} & \multicolumn{1}{c|}{\multirow{2}{*}{\textbf{Mean}}} & \multicolumn{1}{c|}{\multirow{2}{*}{\textbf{Min}}} & \multirow{2}{*}{\textbf{Max}} & \multicolumn{1}{c|}{\multirow{2}{*}{\textbf{Mean}}} & \multicolumn{1}{c|}{\multirow{2}{*}{\textbf{Min}}} & \multirow{2}{*}{\textbf{Max}} & \multicolumn{2}{c|}{\textbf{Telegram}}                                & \multicolumn{2}{c|}{\textbf{Twitch}}             \\ \cline{11-14} 
\multicolumn{1}{|c|}{}                                 & \multicolumn{1}{c|}{}                                & \multicolumn{1}{c|}{}                              &                               & \multicolumn{1}{c|}{}                               & \multicolumn{1}{c|}{}                              &                               & \multicolumn{1}{c|}{}                               & \multicolumn{1}{c|}{}                              &                               & \multicolumn{1}{c|}{\textbf{Pos}} & \multicolumn{1}{c|}{\textbf{Neg}} & \multicolumn{1}{c|}{\textbf{Pos}} & \textbf{Neg} \\ \hline
Train                                                  & \multicolumn{1}{c|}{350}                             & \multicolumn{1}{c|}{172}                           & 178                           & \multicolumn{1}{c|}{64}                             & \multicolumn{1}{c|}{8}                             & 146                           & \multicolumn{1}{c|}{7}                              & \multicolumn{1}{c|}{1}                             & 467                           & \multicolumn{1}{c|}{105}          & \multicolumn{1}{c|}{115}          & \multicolumn{1}{c|}{67}           & 63           \\
Trial                                                  & \multicolumn{1}{c|}{7}                               & \multicolumn{1}{c|}{3}                             & 4                             & \multicolumn{1}{c|}{63}                             & \multicolumn{1}{c|}{1}                             & 135                           & \multicolumn{1}{c|}{5}                              & \multicolumn{1}{c|}{1}                             & 148                           & \multicolumn{1}{c|}{2}            & \multicolumn{1}{c|}{2}            & \multicolumn{1}{c|}{1}            & 2            \\
Test                                                   & \multicolumn{1}{c|}{160}                             & \multicolumn{1}{c|}{83}                            & 77                            & \multicolumn{1}{c|}{59}                             & \multicolumn{1}{c|}{8}                             & 140                           & \multicolumn{1}{c|}{7}                              & \multicolumn{1}{c|}{1}                             & 202                           & \multicolumn{1}{c|}{50}           & \multicolumn{1}{c|}{40}           & \multicolumn{1}{c|}{33}           & 37           \\ \hline
\end{tabular}
\label{tab:datasets}
\end{table}

We conducted a preliminary exploration of the available training sets (357 users across the Train and Trial sets), aiming to analyze the user textual content. First, we calculated the cosine similarity on the TF-IDF representations generated from the complete vocabulary of each class, obtaining a value of 0.854, indicating a high lexical similarity between positive and negative users. Next, we used the Jaccard index to assess the lexical overlap between the classes, considering the 1,000 most frequent words in each. The resulting value was 0.581, corresponding to 735 shared words out of 1,265 unique words, further reinforcing the significant lexical overlap between the two classes. Inspection of these shared words revealed topics such as \emph{cryptocurrencies}, \emph{financial markets}, \emph{games}, \emph{betting}, \emph{digital platforms}, and various \emph{emotional states}. The remaining words unique to each class were strongly linked to these topics, reflecting subtle differences. For instance, positive users tended to use more technical and advanced language, referencing specific platforms (e.g., \emph{BingX}, \emph{Winamax}, \emph{Ledger}, \emph{Discord}, \emph{OKEx}), financial market concepts (e.g., \emph{Elliot}, \emph{scalping}, \emph{velas}—candles, \emph{store}, \emph{\textit{liquidez}}—liquidity), and more active participation in forums and communities. In contrast, negative users displayed less technical language, with expressions suggesting a more cautious attitude (e.g., \emph{aprender}—learn, \emph{consejo}—advice, \emph{ojalá}—hopefully, \emph{imposible}—impossible, \emph{paciencia}—patience), possibly reflecting less experience with these topics. 
As part of the study, we also found that 80\% of users made most of their posts during nighttime hours (between 6 p.m. and 6 a.m.), with this tendency being slightly stronger among positive users (82\%) compared to negative users (78\%). It should be noted that the timestamps are in UTC+1 (+01:00), although it is unclear whether this timezone corresponds to the actual location of each user. After manually inspecting posts, personal contexts, and user dynamics, no clear pattern emerged that could differentiate high-risk from low-risk users. Therefore, this exploration suggests that the task is indeed challenging, which motivated the models proposed by our team.

\section{Methodology}
\label{sec:methodology}

Given the domain complexity and the high lexical similarity between classes, we explored three distinct methods, each aimed at capturing different levels of representation and implementing alternative early classification strategies.
We designed three proposals that combine different classifiers and decision-making strategies in line with the \emph{CPI+DMC approach}.
Below, we describe the models employed and the experiments conducted for their training.

\subsection{Models}

\subsubsection{SS3 Using a Global Decision Policy} 

The SS3 model \cite{burdisso2019text} is a supervised classifier created for ERD problems. It is a robust method that enables incremental user analysis and facilitates the interpretability of the decisions. During the training, SS3 builds a vocabulary with term frequencies for each class. It employs a \emph{global value} ($gv$) function that assigns a score to each term relative to the target classes, considering three parameters: $\sigma$ (smoothness), $\rho$ (sanction), and $\lambda$ (significance). The model aims to emulate human behavior by focusing on key terms when classifying a text, thereby contributing to its interpretability. Internally, it performs a hierarchical analysis at multiple levels (words, sentences, and paragraphs) and applies summary operators to obtain a global value of each. Then, the final classification depends on the sum of the $gv$ scores of all the terms in a user's text. This model constitutes the CPI component of our first proposal, where the representation of samples is created from a frequentist and interpretable model based on term relevance without relying on deep learning. 

To implement the DMC component, we adopted a \emph{global decision policy} previously proposed by our laboratory \cite{loyola2021unsl}. We defined a value \emph{score}$_u$ that estimates a user's overall risk level  based on their post history and the target classes. During user evaluation, we maintain two \emph{confidence values}, $cv_{\text{positive}}$ and $cv_{\text{negative}}$, which accumulate the $gv$ of each observed term across the user's writing history. At post round (\emph{delay}), the user’s current risk level is estimated by normalizing these cumulative values through a \emph{softmax}-based transformation: 
\begin{equation}
\text{\emph{score}}_u =
\text{\emph{softmax}}\left(
\left[
\frac{cv_{\text{positive}}}{\text{\emph{delay}}},
\frac{cv_{\text{negative}}}{\text{\emph{delay}}}
\right]
\right)_{\text{positive}}
\label{eq:ss3-norm}
\end{equation}
This normalization ensures that the resulting scores are within the range [0,1] and allows a fairer comparison between users by mitigating the impact of very short or very long posting histories. In this way, \emph{score$_u$} represents the relative likelihood that the user belongs to the positive class and serves as the basis for the decision-making process that determines whether the user should be classified as positive or negative:

\begin{equation}
\text{\emph{decision}}_u = 
    \begin{cases}
        1,          & \text{if } \text{\emph{score}}_u > \text{median}(\text{\emph{scores}}) + \gamma \cdot \text{MAD}(\emph{scores})\\
        0,          & \text{otherwise}.
    \end{cases}
\label{eq:ss3-policy}
\end{equation}
This policy uses a dynamic threshold defined by the median of all users' scores ($\text{\emph{scores}} = \{\text{\emph{score}}_u|u\in \text{Users}\}$) and the median absolute deviation (MAD), which together define an uncertainty interval: $\text{median}(\text{\emph{scores}}) \pm \gamma \cdot \text{MAD}(\emph{scores})$. The hyperparameter $\gamma$ controls how much a user's $\text{\emph{score}}_u$ must deviate from the median to be classified as positive. In other words, a user is considered at risk if their $\text{\emph{score}}_u$ is significantly higher than most users' scores. 

\subsubsection{Extended BERT Using a History-Based Decision Policy}
We used this transformer-based version as a baseline. It involves fine-tuning a pre-trained BERT model by expanding its original vocabulary, enabling it to represent domain-specific terms previously unknown to the model and whose semantics can contribute to the classification task.  
Specifically, we used BETO \cite{CaneteCFP2020}, a BERT model pre-trained on Spanish corpora.
To identify the new vocabulary terms, we relied on the SS3 model (from our previous proposal) to rank the words according to their relevance to the positive class, from which we selected those to incorporate into the model. In this way, the CPI component of our second proposal aims to obtain distributed and contextualized text representations enriched with domain-relevant terms.

For the DMC component, we employed a decision policy based on the model prediction history during early user detection, referred to as the \emph{history-based rule} \cite{loyola2022unsl, thompson2023strategies, thompson2023early, thompson2024time, thompson2024temporal}. Since transformer models have limitations on the number of tokens they can process, we used a sliding window that concatenates the current post with the previous N ones. At each step, the model predicts the current window and applies the \emph{history-based rule}:
\begin{equation}
\text{\emph{decision}}_u = 
    \begin{cases}
        1,          & \text{if } \sum_{i=1}^{t} \mathbb{I}(p_i \geq \tau) \geq T\\
        0,          & \text{otherwise}.
    \end{cases}
\label{eq:historical-rule}
\end{equation}
Where $p_i$ is the predicted probability at round $i$, $\tau$ is the decision threshold, $T$ is the number of required positive predictions, and $\mathbb{I}(\cdot)$ is the indicator function. Then, $\tau$ and $T$ are hyperparameters that determine the sensitivity and tolerance of the policy, respectively, and were tuned based on the model behavior during the early evaluation of the users. To achieve this, we used the \emph{mock-server} tool\footnote{Available at: \url{https://github.com/jmloyola/erisk_mock_server}}, which simulates an early detection environment through rounds of posts and response submissions, enabling the evaluation of the model's performance through various metrics.

\subsubsection{SBERT Using a History-Based Decision Policy}
The third variant relied on \emph{Sentence-BERT} (SBERT) \cite{reimers-2019-sentence-bert}, a BERT-based model adapted with a Siamese architecture to generate dense, sentence-level representations, capturing semantic relationships to solve tasks, such as classification and semantic search. For the CPI component, we used SetFit (Sentence Transformer Fine-tuning) \cite{tunstall2022efficient}, an efficient framework designed for few-shot scenarios based on contrastive learning. The SBERT encoder is fine-tuned by automatically generating pairs of examples (positive and negative) from the original dataset and training the model to produce embeddings that are closer for examples from the same class and distant from those of different classes. This process results in a discriminative semantic space that supports class separation according to sentence-level representations obtained by the fine-tuned encoder. Then, an independent classifier is trained on the resulting embeddings without further modifying the encoder, leading to an efficient and effective method for classification tasks in complex domains. For DMC, we applied the same \emph{history-based rule} (Equation \ref{eq:historical-rule}), considering the model prediction history to decide when to trigger a risk alert.

\subsection{Experiments}
Following the \emph{CPI+DMC approach}, the experimentation was organized in two stages. In the first stage, we explored different configurations and hyperparameters to find optimal models for user classification. In the second stage, we evaluated different decision-making policies by testing the previously selected models in an early detection environment using the \emph{mock-server} tool. The data usage and specific model configurations adopted by our team are detailed below.

\subsubsection{Data and preprocessing}
We used the textual content of user posts and discarded metadata such as date and platform. 
Although we considered this information, preliminary results showed no performance improvements, and no substantial patterns were found to justify its use.
We merged the Train and Trial sets for the experiments, resulting in 357 samples: 257 for model training and validation and 100 for evaluation in an early detection environment while maintaining a balanced distribution between classes. Preprocessing included converting texts to lowercase, transforming Unicode and HTML sequences into corresponding symbols, normalizing URLs using the 'weblink' token, removing repeated words, and applying other basic text-cleaning operations.

\subsubsection{Model setup}
\begin{itemize}[\indent{}]
\item\textbf{UNSL\#0.} 
SS3 model trained on character trigrams using the hyperparameters $\sigma$=0.44, $\rho$=0.5, and $\lambda$=0.86, selected via grid-search optimized by the F$_1$ metric. A global decision policy was applied, configured with $\gamma$=0.5.

\item\textbf{UNSL\#1.} 
We used the BETO model (checkpoint: 
\href{https://huggingface.co/dccuchile/bert-base-spanish-wwm-uncased}{\emph{dccuchile/bert-base-spanish-wwm-uncased}}) and included 25 domain-relevant words extracted from the SS3 model, considering confidence values assigned to the positive class. This extension allowed us to include terms originally outside the BETO vocabulary, such as \emph{rebote} (rebound), \emph{combi} (combo bets or parlays), \emph{divergencia} (divergence), \emph{BingX}, \emph{scalping}, and \emph{velita} (candlestick), among others. The remaining hyperparameters were: optimizer = \texttt{AdamW}, learning\_rate = \texttt{5E-5}, scheduler = \texttt{LinearSchedulerWarmup}, batch\_size = 32, and n\_epochs = 10.
The checkpoint with the highest F$_1$ score on the validation set was selected.
We used a \emph{history-based rule} configured with $T$=10 and $\tau$=0.6. 
\item\textbf{UNSL\#2.} SBERT model, based on BETO and pre-trained on semantic similarity tasks in Spanish (checkpoint: 
\href{https://huggingface.co/hiiamsid/sentence_similarity_spanish_es}{\emph{hiiamsid/sentence\_similarity\_spanish\_es}}). We fine-tuned the encoder using the \emph{CosineSimilarityLoss} function, followed by a logistic regression classifier trained on the resulting embeddings. The configuration included batch\_size = 16, num\_epochs = 1, num\_iteration = 20, and learning\_rate = \texttt{2E-5}. Finally, we defined a \emph{history-based rule} configured with $T$=10 and $\tau$=0.7.

\end{itemize}

\section{Results}
\label{sec:results}
A total of 38 proposals were submitted by thirteen teams for Task 1. The Organizers evaluated the models using classification and latency metrics, and published an official ranking based on the Macro F$_1$. Table \ref{tab:results} summarizes the results obtained by our models and compares them with some of the most relevant proposals (complete official results reported in \cite{MentalRiskES2025}). We highlight the following observations:

\begin{itemize}[]
\item   
UNSL\#2 reached first place in the overall ranking, achieving a Macro F$_1$ of 0.567. It also obtained the second-best results in Accuracy, Macro Recall, Micro Precision, Micro Recall, and Micro F$_1$. Additionally, it showed acceptable F$_{latency}$, considering the top three models.

\item 
UNSL\#0 was ranked second, with a Macro F$_1$ of 0.563, and delivered comparable performance to UNSL\#2. It excelled in several metrics, obtaining the best scores in Accuracy, Macro Recall, Micro Precision, Micro Recall, and Micro F$_1$, as well as the best F$_{latency}$ among the top-ranked models. It also achieved an ERDE$_{30}$ of 0.284, better than the overall average (0.325) and comparable with the best, PLN-PPM-ISB\#0.

\item 
UNSL\#1 model achieved 16th with a Macro F$_1$ of 0.444, outperforming both the mean (0.426) and the median (0.429) of all submissions, and providing an acceptable baseline for comparison.

\item 
Among other proposals, I2C-UHU-Rigel\#1 achieved the third-best Macro F$_1$, VerbaNexAI-Lab\#0 (27th) the best ERDE$_{5}$, PLN-PPM-ISB\#0 (17th) the best ERDE$_{30}$ and the second-best F$_{latency}$, while Robertuito (20th) obtained the best Macro Precision and F$_{latency}$.

\end{itemize}

\begin{table}[ht!]
\caption{Decision-based evaluation results for Task 1 in MentalRiskES 2025. The best three models, ranked by the Organizers based on the Macro F$_1$ metric, are shown. Additionally, the highest performances in ERDE$_{5}$, ERDE$_{30}$, and F$_{latency}$ are included, along with the Mean and Median across all proposals. Values in bold and underlined indicate the first and second-best scores, respectively.}
\resizebox{\textwidth}{!}{%
\begin{tabular}{|clccc
>{\columncolor[HTML]{D9D9D9}}c cccccc|}
\hline
\multicolumn{1}{|l|}{\textbf{Rank}} & \textbf{Model\#Run} & \textbf{Acc} & \textbf{\begin{tabular}[c]{@{}c@{}}Macro\\ P\end{tabular}} & \textbf{\begin{tabular}[c]{@{}c@{}}Macro\\ R\end{tabular}} & \textbf{\begin{tabular}[c]{@{}c@{}}Macro\\ F$_{1}$\end{tabular}} & \textbf{\begin{tabular}[c]{@{}c@{}}Micro\\ P\end{tabular}} & \textbf{\begin{tabular}[c]{@{}c@{}}Micro\\ R\end{tabular}} & \textbf{\begin{tabular}[c]{@{}c@{}}Micro\\ F$_{1}$\end{tabular}} & \textbf{ERDE$_{5}\downarrow$} & \textbf{ERDE$_{30}\downarrow$} & \textbf{F$_{latency}$} \\ \hline
\multicolumn{1}{|c|}{1} & UNSL\#2 & {\ul 0.569} & 0.568 & {\ul 0.567} & \textbf{0.567} & {\ul 0.569} & {\ul 0.569} & {\ul 0.569} & 0.639 & 0.389 & 0.506 \\
\multicolumn{1}{|c|}{2} & UNSL\#0 & \textbf{0.581} & 0.586 & \textbf{0.574} & {\ul 0.563} & \textbf{0.581} & \textbf{0.581} & \textbf{0.581} & 0.515 & 0.284 & 0.628 \\
\multicolumn{1}{|c|}{3} & I2C-UHU-Rigel\#1 & 0.556 & 0.555 & 0.553 & 0.551 & 0.556 & 0.556 & 0.556 & 0.600 & 0.432 & 0.496 \\ \hline
\multicolumn{1}{|c|}{16} & UNSL\#1 & 0.475 & 0.459 & 0.467 & 0.444 & \cellcolor[HTML]{FFFFFF}0.475 & 0.475 & 0.475 & 0.707 & 0.476 & 0.467 \\
\multicolumn{1}{|c|}{17} & PLN-PPM-ISB\#0 & 0.550 & 0.632 & 0.534 & 0.436 & 0.550 & 0.550 & 0.550 & 0.316 & \textbf{0.242} & {\ul 0.683} \\
\multicolumn{1}{|c|}{20} & Robertuito & 0.550 & \textbf{0.657} & 0.533 & 0.428 & 0.550 & 0.550 & 0.550 & 0.329 & 0.252 & \textbf{0.685} \\
\multicolumn{1}{|c|}{27} & VerbaNexAI-Lab\#0 & 0.519 & 0.259 & 0.500 & 0.342 & 0.519 & 0.519 & 0.519 & \textbf{0.274} & 0.250 & 0.677 \\ \hline
\multicolumn{2}{|c}{\textit{Mean}} & 0.524 & 0.457 & 0.512 & 0.426 & 0.524 & 0.524 & 0.524 & 0.473 & 0.325 & 0.587 \\
\multicolumn{2}{|c}{\textit{Median}} & 0.519 & 0.490 & 0.500 & 0.429 & 0.519 & 0.519 & 0.519 & 0.445 & 0.283 & 0.657 \\ \hline
\end{tabular}%
}
\label{tab:results}
\end{table}

In addition, the Organizers requested that the teams provide estimates of energy consumption and resource usage to assess the computational and environmental impact using the CodeCarbon library\footnote{Available at: \url{https://github.com/mlco2/codecarbon}}.
As shown in Table \ref{tab:emissions}, our models were executed on the same hardware configuration, with each inference taking an average of 2.3 seconds, consuming approximately 7E-05 kWh (kilowatt-hour) of energy, and producing 1.66E-07 kgCO$_2$eq (kilograms of carbon dioxide equivalent), all values significantly lower than the recorded mean.

\begin{table}[ht!]
\caption{Efficiency and resource usage metrics for our team’s submissions in Task 1. Mean values across all proposals are reported when applicable.}
\resizebox{\textwidth}{!}{%
\begin{tabular}{|ccccccccccrr|}
\hline
\multicolumn{1}{|l}{\textbf{Team\#Run}} & \textbf{\begin{tabular}[c]{@{}c@{}}Duration\\ (sec)\end{tabular}} & \textbf{Emissions} & \textbf{\begin{tabular}[c]{@{}c@{}}CPU\\ energy\end{tabular}} & \textbf{\begin{tabular}[c]{@{}c@{}}GPU\\ energy\end{tabular}} & \textbf{\begin{tabular}[c]{@{}c@{}}RAM\\ energy\end{tabular}} & \textbf{\begin{tabular}[c]{@{}c@{}}Consumed\\ energy\end{tabular}} & \textbf{\begin{tabular}[c]{@{}c@{}}CPU\\ count\end{tabular}} & \textbf{\begin{tabular}[c]{@{}c@{}}GPU\\ count\end{tabular}} & \textbf{\begin{tabular}[c]{@{}c@{}}Total RAM\\size\end{tabular}} & \multicolumn{1}{c}{\textbf{\begin{tabular}[c]{@{}c@{}}CPU\\ model\end{tabular}}} & \multicolumn{1}{c|}{\textbf{\begin{tabular}[c]{@{}c@{}}GPU\\ model\end{tabular}}} \\ \hline
\textbf{UNSL\#0} & 2.27E+00 & 1.66E-07 & 2.68E-05 & 4.29E-05 & 2.88E-07 & 6.99E-05 & 32 & 1 & 63.730 & \multirow{4}{*}{\begin{tabular}[c]{@{}r@{}}Intel(R) \\ Core(TM) i9-14900HX\end{tabular}} & \multirow{4}{*}{\begin{tabular}[c]{@{}r@{}}1 x NVIDIA GeForce \\ RTX 4090 Laptop GPU\end{tabular}} \\
\multirow{2}{*}{\textbf{UNSL\#1}} & \multirow{2}{*}{2.27E+00} & \multirow{2}{*}{1.66E-07} & \multirow{2}{*}{2.68E-05} & \multirow{2}{*}{4.29E-05} & \multirow{2}{*}{2.88E-07} & \multirow{2}{*}{6.99E-05} & \multirow{2}{*}{32} & \multirow{2}{*}{1} & \multirow{2}{*}{63.730} &  &  \\
 &  &  &  &  &  &  &  &  &  &  &  \\
\textbf{UNSL\#2} & 2.27E+00 & 1.66E-07 & 2.68E-05 & 4.29E-05 & 2.88E-07 & 6.99E-05 & 32 & 1 & 63.730 &  &  \\
\hline
\rule{0pt}{0.8em}
\textit{Mean} & 3.03E+01 & 6.59E-04 & 6.01E-04 & 5.98E-04 & 9.45E-05 & 1.30E-03 & 16 & 1 & 40.27 & \multicolumn{1}{c}{-} & \multicolumn{1}{c|}{-} \\ \hline
\end{tabular}%
}
\label{tab:emissions}
\end{table}

\subsection{Error analysis}
\begin{figure}[h]
\centering
\includegraphics[width=0.8\textwidth]{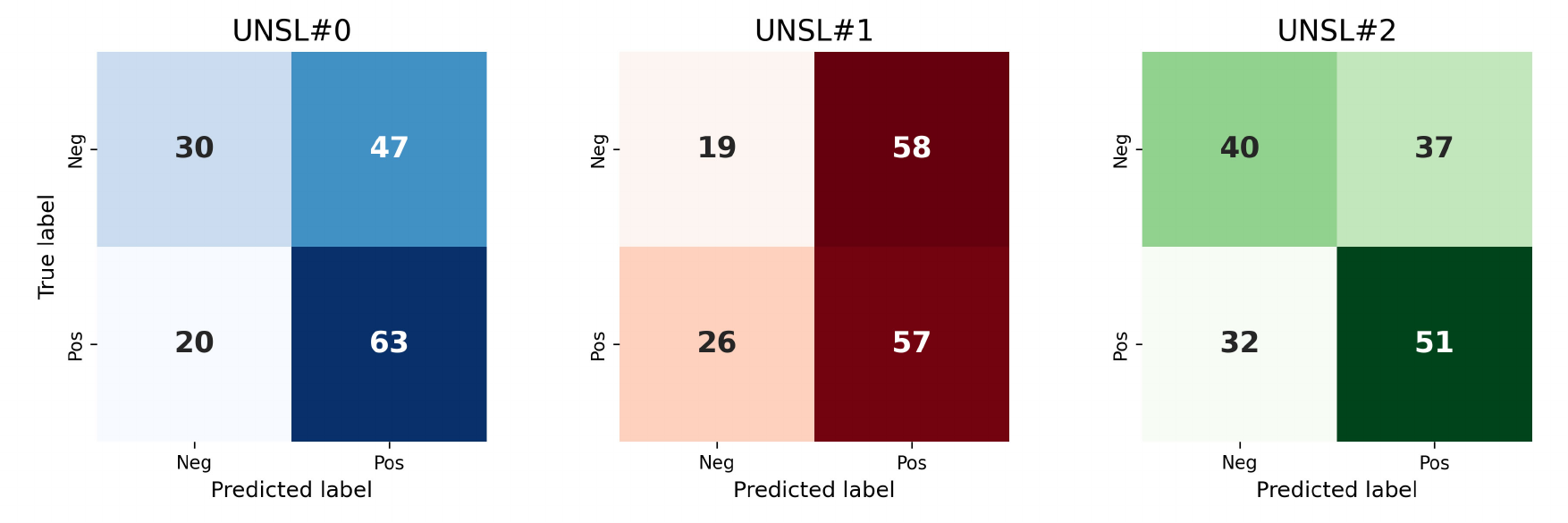} 
\caption{Confusion matrices for the three models proposed in Task 1.}
\label{fig:cmatrix}
\end{figure}

Figure \ref{fig:cmatrix} presents the confusion matrix for each proposal, based on the first time each model predicted a user as positive during its post-by-post interaction with the server. Among the three models, UNSL\#0 detected the most true positives (TPs), while UNSL\#2 identified the most true negatives (TNs). UNSL\#2 also had the lowest number of false positives (FPs) and a similar number of false negatives (FNs), achieving a balance between precision and recall, suggesting a more conservative approach to detecting positive cases. In contrast, UNSL\#0 reduced FNs but increased FPs, reflecting a more sensitive yet less precise strategy that prioritizes early detection, even at the expense of generating more incorrect alerts. Meanwhile, UNSL\#1 exhibited intermediate performance in FNs, the highest number of FPs, and the lowest TNs, indicating difficulties distinguishing between the two classes.

\begin{figure}[h]
\centering
\includegraphics[width=1\textwidth]{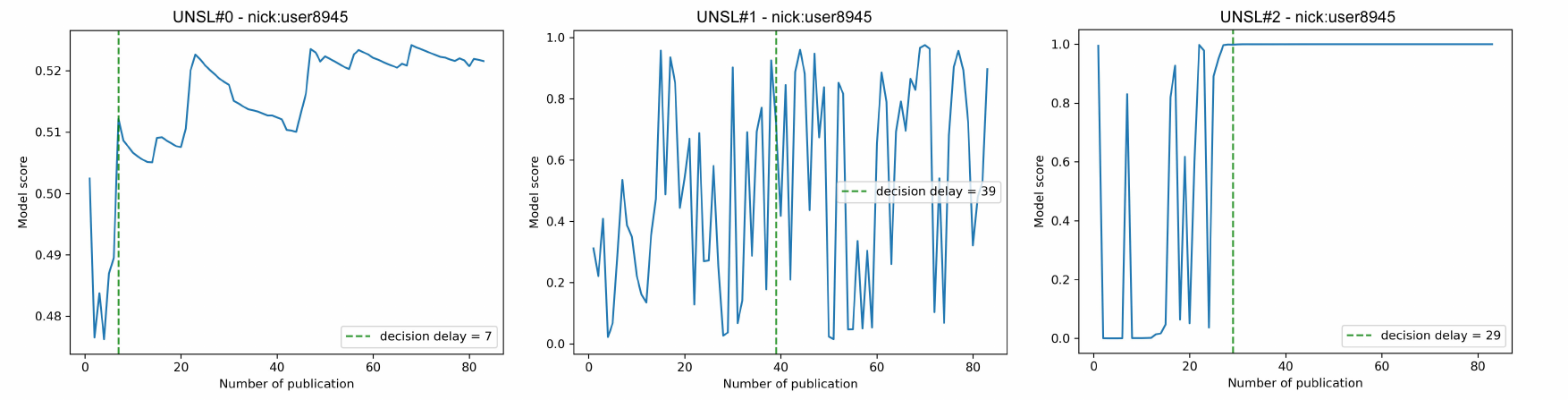} 
\caption{Comparison of UNSL\#0, UNSL\#1, and UNSL\#2 on a positive user from Task 1. The $x$-axis shows the rounds of posts (number of publication), while the $y$-axis indicates the model's predicted score at each decision point. Dashed green lines mark the moment when each model made its final correct prediction.}
\label{fig:model_decisions}
\end{figure}

Figure \ref{fig:model_decisions} shows the predictions of the three models when analyzing a positive user from Task 1. Despite following different strategies, all three models consistently showed signs that the user was at high risk throughout the analysis. UNSL\#0 predicted the user as positive at round 7, UNSL\#1 at round 39, and UNSL\#2 at round 29. UNSL\#0 exhibits fewer score variations, with confidence values remaining close to 0.5, and issued an alert after exceeding the uncertainty interval defined by the decision policy. UNSL\#1 displays higher variability, with predictions oscillating between high and low probabilities across many rounds. In contrast, UNSL\#2 initially presents isolated high scores that later stabilize toward the end of the analysis. Both UNSL\#1 and UNSL\#2 benefited from the history-based rule, which allowed them to tolerate fluctuations and wait for consistent signals before issuing a final decision.

\begin{figure}[h]
\centering
\includegraphics[width=0.50\textwidth]{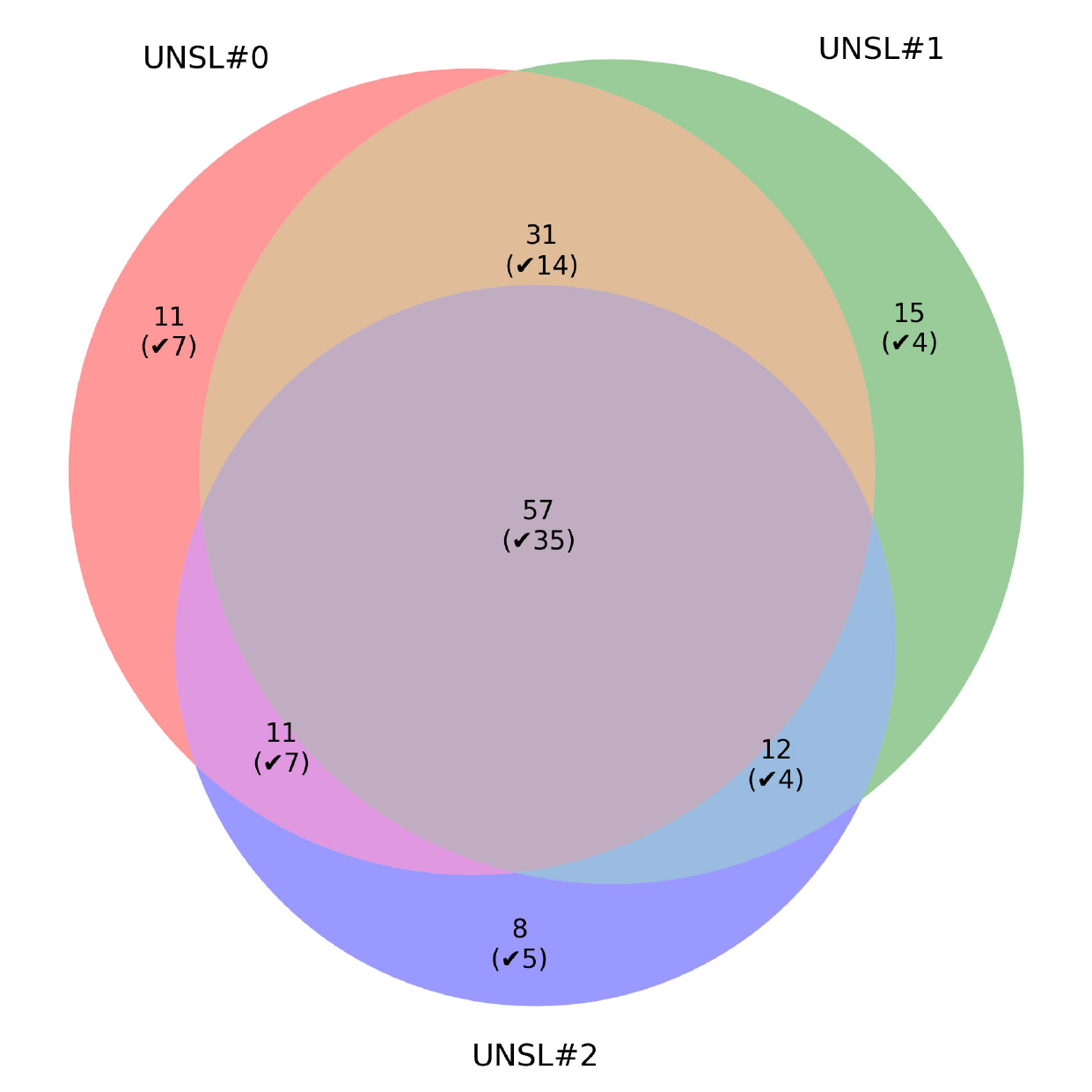} 
\caption{Venn diagram comparing the users predicted as positive by the three models proposed in Task 1. The overlaps and values between sets illustrate the agreement among models, with the number of correct predictions shown in parentheses.}
\label{fig:model_agreements}
\end{figure}

The Venn diagram in Figure \ref{fig:model_agreements} illustrates the distribution of positive predictions across the three models, highlighting their overlaps and divergences. The three models agreed on 57 positive instances (35 correctly classified), while the remaining 22 were FPs shared by all three. Upon examining these presumably incorrect samples, we observed recurring themes such as \emph{sports betting}, \emph{video games with elements of chance}, and \emph{cryptocurrency trading}. These cases often featured behaviors such as \emph{financial speculation}, \emph{active engagement in games}, and \emph{intense emotional expressions related to wins and losses}. Only a minority consisted of short or ambiguous messages that required deeper contextual or linguistic interpretation. 
This suggests that many of these FPs may be due to the limitations of the corpus in clearly distinguishing between risk levels.
The analysis highlights the intrinsic complexity of the task, where ambiguity and overlap between users at high and low risk, not only at the lexical level (as noted in Section \ref{sec:corpus}) but also semantic, can lead to misclassifications by both predictive models and human evaluators.

To assess the semantic consistency of the predictions made by the UNSL\#2 and UNSL\#0 models, we show some illustrative examples using the following sentences:
\begin{quote}
\text{S1}: \emph{``hoy jugué durante horas en BingX buscando ese rebote que me haría recuperar lo que perdí ayer... pero me comí terrible divergencia!''}
(I played for hours today on BingX looking for that rebound that would make up for what I lost yesterday... but I got caught with a terrible divergence!).
\end{quote}
\begin{quote}
\text{S2}: \emph{``tarde o temprano llegará el gool, pero espero que no sea en el primer tiempo porque entre con bastante''}
(Sooner or later the goal will come, but I hope it's not in the first half because I'm going in with a lot).
\end{quote}
\begin{quote}
\text{S3}: \emph{``he hecho algunas pequeñas inversiones, pero la verdad que no me gustan mucho estas cosas''}
(I've made some small investments, but I'm not really into this kind of thing).
\end{quote}
The UNSL\#2 model classified the first two sentences as positive and the third as negative. To assess the semantic consistency of these predictions, we obtained the embeddings of each sentence using the UNSL\#2 encoder and calculated the cosine similarity between them. We observed that S1 and S2 exhibited high similarity (0.7282), while S3 showed low similarity with both S1 (0.1202) and S2 (0.0065). 
This suggests that the model may be constructing a representation space where sentences related to users with gambling tendencies (e.g., impulsive investments or sports betting) tend to be closer together, while those without these characteristics are farther apart. Since distinguishing between users at high and low risk can be ambiguous and challenging, even semantically consistent representations can result in errors if the corpus does not adequately reflect these differences. However, the overall performance of UNSL\#2 in detecting the positive class depends on both the learned representation space and the classifier that utilizes these representations for prediction.

\begin{figure}[h]
\centering
\includegraphics[width=0.95\textwidth]{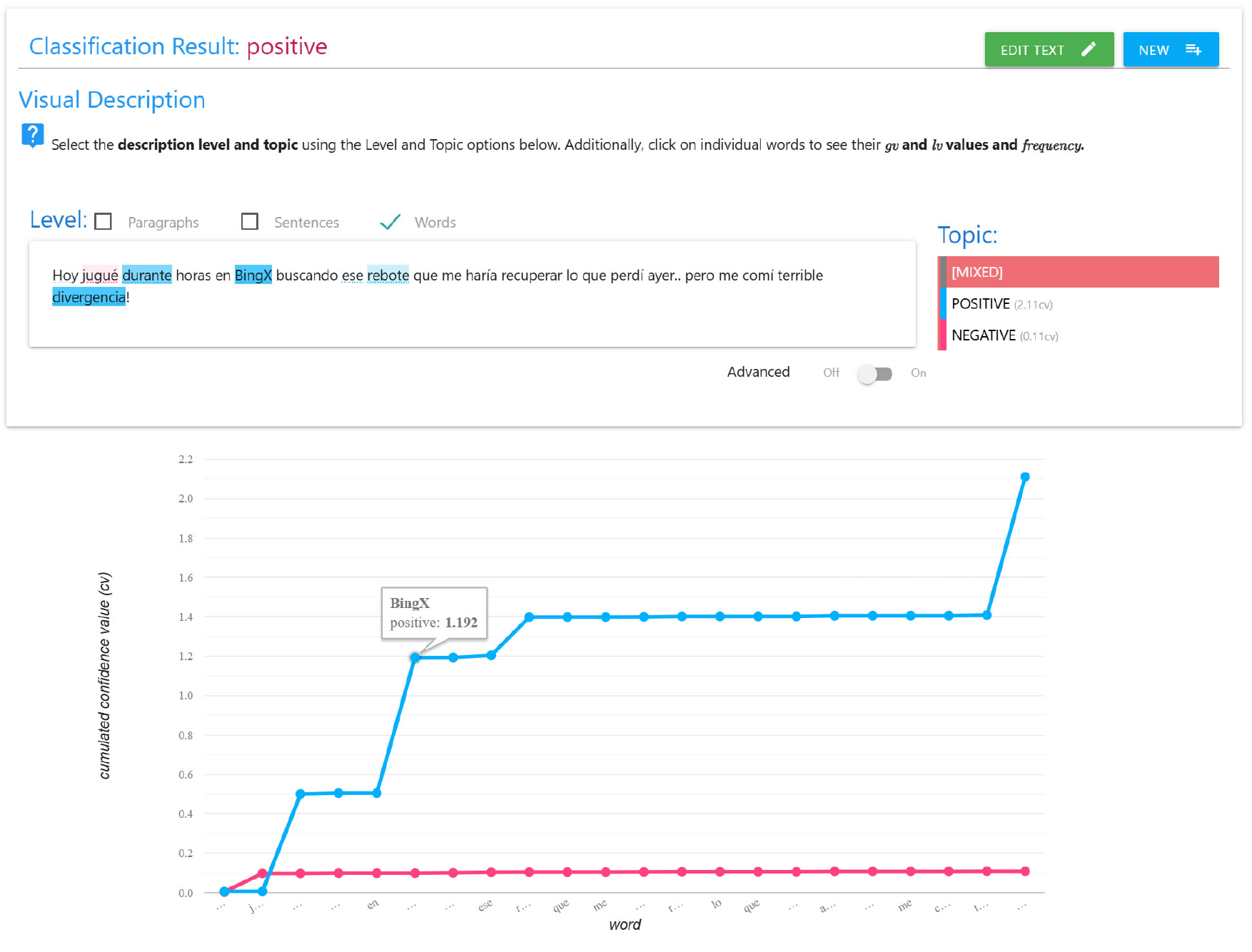} 
\caption{Visual explanation of the evaluation process used by UNSL\#0. The model has a visualization tool to improve interpretability by using the cumulated confidence values ($cv$), associated with each word in the sentence.  Words are color-coded according to their contribution to the positive class. As the analysis progresses, the $cv$ value increases, with certain words such \emph{BingX}, \emph{rebote} (bounce), and \emph{divergencia} (divergence) progressively reinforcing the model’s tendency to classify the sentence as positive. }
\label{fig:ss3_decs}
\end{figure}

The UNSL\#0 model, based on SS3, facilitates interpretation by providing a tool to visualize the information it considers relevant for each prediction. For instance, Figure \ref{fig:ss3_decs} shows that the model classified sentence \text{S1} as positive, assigning distinct relevance scores to terms such as \emph{durante} (during), \emph{BingX}, \emph{rebote} (bounce), and \emph{divergencia} (divergence) for the positive class, while the term \emph{jugué} (I played) was slightly associated with the negative class. Additionally, the cumulative $cv$ increases as the sentence progresses, especially after encountering the word \emph{BingX}, which strongly contributed to the model’s decision. 

\subsection{Final Considerations}
The results highlight the inherent complexity of the task, particularly the challenge of distinguishing between positive and negative classes. In \cite{thompson2025hacia}, we explored Large Language Models (LLMs) to address the ERD in depression by introducing a reasoning criterion grounded in expert knowledge. The goal was not only to detect positive cases, but also to generate explanations that justify the model's decisions. While defining such criteria is difficult, this strategy can enhance data construction and interpretation, allowing for more precise identification of the specific moments when risk signals emerge. It may also contribute to more reliable evaluation metrics, such as an adaptive version of ERDE$\theta$, where each user has an individual threshold $\theta$ based on their behavior, penalizing delayed decisions more fairly.

On the other hand, the ERDE metric was defined in \cite{losada2016test} to penalize delayed TPs, while the cost associated with FPs and FNs depends on the domain. The authors also point out that negative cases correspond to non-risk situations, in which early or urgent intervention is not required. However, in our case, negative users are already at some level of risk, which changes the notion of \emph{``early detection''}: it is no longer about identifying an early risk case under the assumption that risk is initially absent, but rather about recognizing the moment when a user crosses a critical threshold that justifies more serious concern. This shift in perspective may help explain why most participants underperformed on early detection metrics, particularly ERDE$\theta$. Considering that in Task 1 the average number of posts per user was around 60 (see Table \ref{tab:datasets}), a higher threshold (e.g., $\theta$ = 50) could have helped avoid severe penalties during the initial stages of analysis, when much of the evidence was still unavailable.

This year’s edition raises relevant conceptual challenges for ERD, including how risk is defined, how it is measured, and what types of decisions we expect models to make. For instance, an FP made with very little evidence might represent a more serious false alarm than a late FP, in which the user already shows ambiguous patterns, something even possibly acceptable from a preventive perspective. The same holds for FNs: delaying a prediction may be reasonable when signals are weak, but as more information accumulates over time, the model should be able to detect evident signs of risk.

\section{Conclusion}
\label{sec:conclusion}
Our laboratory solved Task 1 of MentalRiskES 2025 by presenting three proposals based on the \emph{CPI+DMC approach}. Two of these proposals achieved outstanding results among all participating teams, demonstrating that ERD can be addressed by balancing classification performance and decision-making speed through a modular and independent approach. Corpus exploration played a crucial role in the selection of the methods used. The results highlighted the complexity of distinguishing between high-risk and low-risk users, which can be challenging even from a human perspective.
It is essential to continue researching strategies that enhance the quality and interpretation of data, particularly in the ERD of mental health, where transparent and reliable systems are needed to support identification and analysis in critical areas, such as gambling disorder. Furthermore, we will continue exploring new approaches that address ERD by combining predictive effectiveness and speed as a single combined objective.

\begin{acknowledgments}
This work is part of the doctoral research of Horacio Thompson, carried out at the Laboratorio de Investigación y Desarrollo en Inteligencia Computacional (LIDIC), under the project PROICO 03-0620, Argentina. \end{acknowledgments}

\section*{Declaration on Generative AI}
The authors have not employed any Generative AI tools.

\bibliography{cites}

\end{document}